# Single-fingered reconfigurable robotic gripper with a folding mechanism for narrow working spaces

Toshihiro Nishimura[1], *Member, IEEE*, Tsubasa Muryoe[2], Yoshitatsu Asama[3], Hiroki Ikeuchi[3], Ryo Toshima[3], and Tetsuyou Watanabe[1], *Member, IEEE*

*Abstract*—This paper proposes a novel single-fingered reconfigurable robotic gripper for grasping objects in narrow working spaces. The finger of the developed gripper realizes two configurations, namely, the insertion and grasping modes, using only a single motor. In the insertion mode, the finger assumes a thin shape such that it can insert its tip into a narrow space. The grasping mode of the finger is activated through a folding mechanism. Mode switching can be achieved in two ways: switching the mode actively by a motor, or combining passive rotation of the fingertip through contact with the support surface and active motorized construction of the claw. The latter approach is effective when it is unclear how much finger insertion is required for a specific task. The structure provides a simple control scheme. The performance of the proposed robotic gripper design and control methodology was experimentally evaluated. The minimum width of the insertion space required to grasp an object is 4 mm (1 mm, when using a strategy).

## I. INTRODUCTION

THIS paper presents a novel thin-shaped reconfigurable robotic finger for narrow working spaces. The demand for handling objects in narrow spaces, such as picking up an object from a box, is high in realistic situations. Insertion of a finger into a narrow space and grasping capabilities are required for such tasks. A thin-finger design is preferable for insertion. The space occupied by the robotic gripper should be minimized to ensure that it can use as large a workspace as possible within a limited workspace. Thus, the gripper should have as few fingers as possible. This study aimed to develop a single-fingered reconfigurable robotic gripper with a thin width that can operate in a narrow space. A folding mechanism is installed on the robotic finger such that an object can be grasped using a single finger. The basic behavior of the developed gripper is illustrated in Fig. 1(a). The operations utilizing the finger are shown in Figs. 1(b) and 1(c). The finger assumes a thin shape for in insertion mode. After insertion, the links comprising the finger are folded, and the claws of the mini-gripper emerge on the surface of the robotic finger. Thus, the grasping mode has been activated. The object is grasped and picked using the emerged claws. The mode switching and grasping operations are performed using a single motor. Two different methods for activating mode switching have been proposed. The main difference between these methods lies in the approach to activating the fingertip rotation, that is, the lower claw, as shown in Fig. 1(a). If the motor is actively turned on and rotated, the finger structure switches from insertion to grasping mode. This switching is referred to as active switching and is independent of environmental conditions, as shown in Fig. 1(b). If the motor is turned off and the fingertip is inserted

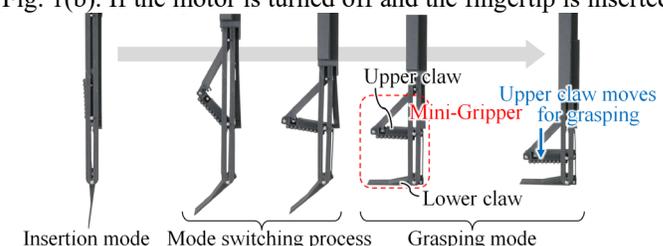
(a) Basic behavior of developed reconfigurable robotic finger

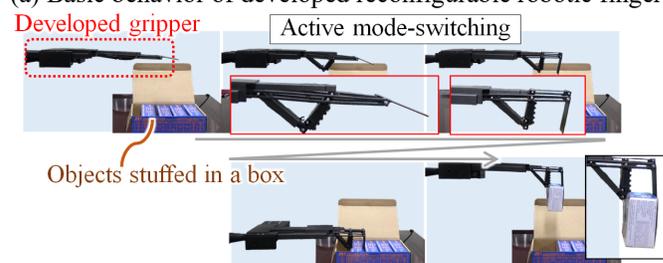
(b) Picking an object stuffed in a box using active switching

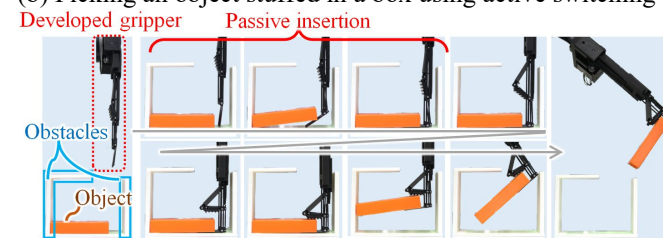
(c) Picking on an object in a narrow space partially enclosed by the top and sides using partial passive switching

Fig. 1. Basic behavior of the developed reconfigurable robotic gripper and object-grasping in narrow spaces

into the surroundings through the movement of the gripper body, the fingertip passively rotates, as it adapts its posture to

Manuscript received: February 24, 2022; Revised: May 25, 2022; Accepted June 28, 2022. This letter was recommended for publication by Associate Editor and Editor Liu Hong upon evaluation of the reviewers' comments. This work was supported by Panasonic Corporation. (*Corresponding authors: Tetsuyou Watanabe; Toshihiro Nishimura.*)

[1]T. Nishimura and T. Watanabe are with the Faculty of Frontier Engineering, Institute of Science and Engineering, Kanazawa University, Kakuma-machi, Kanazawa, 9201192 Japan (e-mail: tnishimura@se.kanazawa-u.ac.jp, te-watanabe@ieee.org).
[2]T. Muryoe is with the Graduated school of Natural science and Technology, Kanazawa University, Kakuma-machi, Kanazawa, 9201192 Japan.
[3]Y. Asama, H. Ikeuchi, and R. Toshima are with the Panasonic Corporation.

Digital Object Identifier (DOI): see top of this page.

the surroundings. Therefore, the fingertip can be inserted under an object on a table, and the object can be scooped in a narrow space by simply moving the gripper body. Mode switching is performed through self-adaptive motion, followed by the active activation of the upper claw, as shown in Fig. 1(c), using motor drives. This switching is referred to as partial passive switching.

*A. Related works*

Grasping objects in narrow spaces is an important task that expands the adoptable area of robots [1][2]. Several studies have attempted to realize this task through the development of end-effectors, such as robotic grippers or wrists. Hasegawa et al. proposed a robotic gripper with a suction mechanism to grasp objects in a narrow space [3]. Hernandez et al. proposed a gripper with pinch-and-suction mechanisms [4]. Fukui et al. developed a robotic hand with thin nails to grasp objects in a box [5]. Negrello et al. designed a robotic wrist to expand the operating range of a manipulator for a task within a narrow space [6]. Littlefield compared the performance of suction and three-fingered grippers for grasping tasks in a narrow space [7]. In [3], [4], and [7], a suction gripper was used; however, it was difficult to grasp objects with rough surfaces stably using suction.

In addition, several robotic grippers with thin fingers, such as those described in [5], have been developed. Wang et al. designed a pneumatic gripper with needles for handling food products [8]. Endo et al. proposed a thin-fingered gripper for grasping food [9]. Jain et al. developed a soft robotic gripper using a thin retractable nail [10]. Morino et al. designed a gripper with an object-pull-in function [11]. However, these thin-fingered robotic grippers were limited by their large sizes in narrow spaces because only the fingertip areas were designed to operate in such spaces.

The use of a single actuator for mode switching is a simple and effective control method for object handling. Belter et al. developed a one-actuator robotic hand with four grasping modes [12]. Liu et al. designed a robotic gripper that can switch between grasping and roll manipulation modes [13]. In [14] and [15], grippers that switch between grasping and object pull-in modes using a single motor were proposed. In addition, our research group developed a gripper with multiple grasping modes [16]–[18]. However, robots that perform tasks in narrow spaces were not considered in these studies.

Several robotic grippers contain a folding structure. In [19]–[21], a gripper inspired by the origami structure was developed in [19]–[21]. Lu et al. developed a gripper with a variable friction surface using a folding mechanism [22]. Orlofsky developed a miniature gripper using an origami structure [21]. An easily exchangeable paper gripper was designed in [23] and [24].

Although several robotic grippers for narrow-space tasks and thin-fingered grippers have been proposed, no attempt has been made to develop a single-fingered reconfigurable robotic gripper with a single actuator for grasping objects in a narrow space. The thin fingers function in the insertion and grasping modes through the folding mechanism. Another important feature is that both active and passive fingertip movement methods can be used in insertion mode.

## II. RECONFIGURABLE ROBOTIC FINGER DESIGN

*A. Functional requirements*

The functional requirements of the developed reconfigurable robotic finger are as follows: 1) compactness, achieved by realizing the insertion and grasping modes using a single actuator, and 2) a thickness of less than 10-mm, less than the thickness of a human finger [25], in the insertion mode.

*B. Structure*

Fig. 2 shows the three-dimensional computer-aided design (3D-CAD) model and two methodologies for the mode switching of the proposed gripper. The gripper comprises a fixing base (light blue part in Fig. 2(a)), robot mounter (black part), motor with pinion gear (black part), slide base (yellow/purple part), four types of links, slide pin, and five connect pins. The four links are upper (pink), middle (red), lower (orange), and fingertip (green) links, connected in series with a slide base via connect pins. The fingertip link of the proposed gripper is 1.0-mm thick; hence, it can insert its finger into a narrow space. The slide base incorporates a rack gear that meshes the pinion gear rotated by the motor and moves relative to the fixing base through motor rotation. The finger design has several key structures for switching between the insertion and grasping modes. The grasping operation of the gripper is realized using four links. The lower link incorporates a slotted hole with protrusions, as shown in Fig. 2(a), to control the behavior of the four links to enable the switch from insertion to grasping mode and vice versa. The fixing base includes guide rails with a slotted hole and a back cover (purple area in Fig. 2(a)). The emergence of the upper claw on the finger, as shown in Fig. 1(a), is facilitated using the upper and middle links. To maintain the configuration of the upper claw following emergence, a lock mechanism with a spring clip in the upper link and clip pin in the middle link is installed, as shown in Figs. 2(b) and 2(c). The stoppers, which prevent singular configurations that can obstruct mode switching, are installed between the upper and middle links and between the lower and fingertip links, as shown in Fig. 2(b).

Two methodologies of switching are utilized, depending on the situation: 1) active switching whereby the mode is switched solely by driving the motor, and 2) combination of passive rotation of the fingertip links through contact with a support surface and active motorized activation of the upper claw. The latter strategy is effective when the required amount of fingertip insertion is unclear. First, the former methodology is presented. Fig. 2(b) shows the mode-switching mechanism accomplished through unidirectional motor rotation. The behavior of the series-connected links when the slide base is moved downward by motor rotation through the rack-and-pinion mechanism is shown in Fig. 2(b). Initially, the slide pin installed on the middle link (Fig. 2(a)) is translated along the slotted hole of the guide rail. The "translation" transforms the configurations of the upper and middle links, pushing the lower link backward. The fingertip link rotates in conjunction with the movement of the lower link. Subsequently, when the slide pin reaches the position where the middle link becomes perpendicular to the guide rail, the pin translation ceases, owing to the protrusion at the slotted hole of the lower link, and the spring clip of the upper link snaps with the clip pin of the middle link. These steps trigger the emergence of the upper claw on the finger through the upper and middle links. The fingertip link rotates 90°, following which the lower claw emerges. If the slide base continues to move, the connect pin at the lower link deforms the lower link, climbs over the protrusion, and eventually moves below the protruding position. Correspondingly, the emerged upper claw moves below the protrusion. The back cover constrains the translation of the lower link, and the posture of the fingertip link is maintained in the grasping mode. After mode switching, open/closed motion for grasping objects is achieved by the movement of the upper claw through motor control.

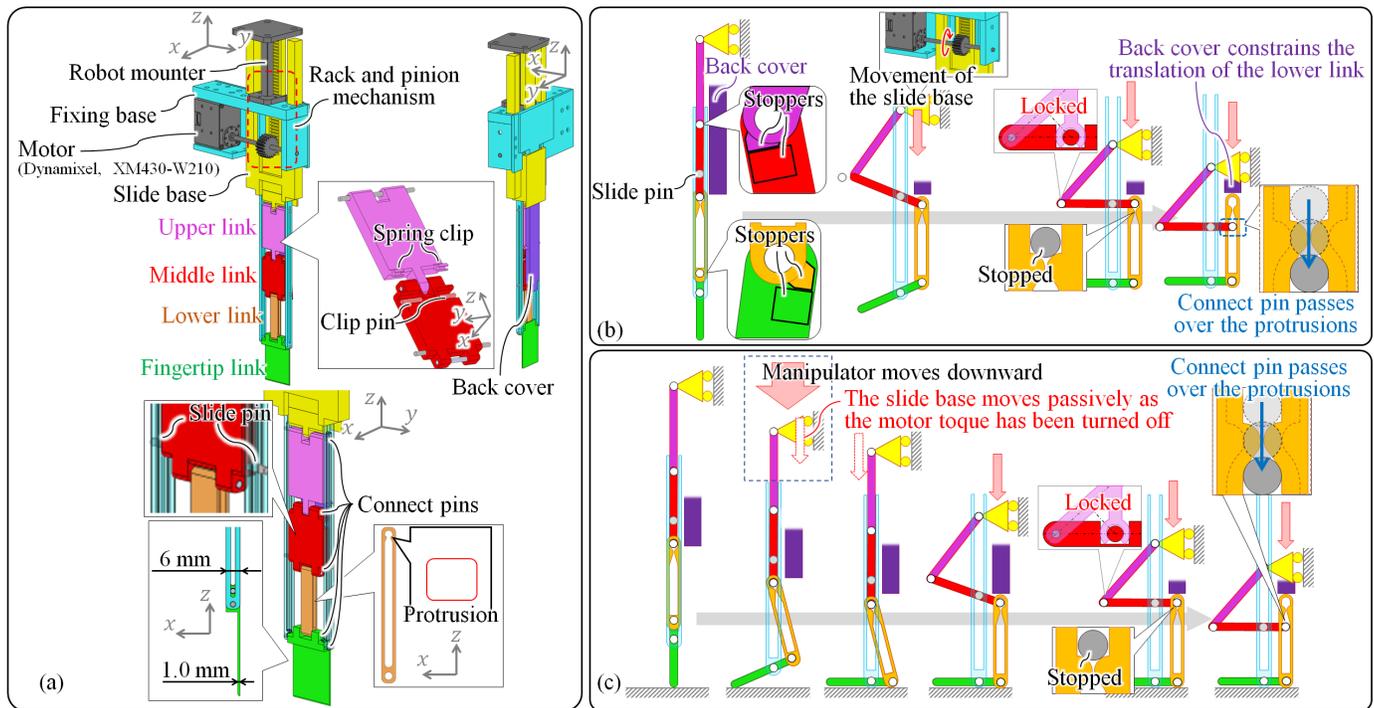

**Fig. 2.** Structure of the developed gripper and procedures for switching from the insertion mode to the grasping mode by two methodologies: (a) 3D-CAD model, (b) switching procedure by active motor rotation only, and (c) switching procedure while using a contact with a supporting surface

Next, a switching procedure that uses contact with the supporting surface was presented. In a narrow space, the space for fingertip motion is limited; thus, it is effective to insert the fingertip under the object to scoop and grasp it. Although this insertion can be performed by active motor driving, it is necessary to sense, recognize, and control the fingertip. However, this strategy is difficult to implement in narrow spaces. Hence, the finger is designed such that the fingertip link rotates passively upon contact with a supporting surface to enable insertion of the fingertip under the object by simply moving the gripper body downward. The procedure for switching the motion mode using contact with a supporting surface is shown in Fig. 2(c). First, the motor torque is turned off. Then, the manipulator attached to the gripper moves downwards. After the fingertip link establishes contact with the supporting surface, a contact force is transmitted to the motor through the series-connected links. The motor is then rotated passively, and the slide base moves downward passively. The movement of the slide base passively rotates the fingertip link, which facilitates the insertion of the fingertip link under an object. Subsequently, the motor is turned on and rotated actively to move the slide base downward. The movement of the slide base triggers the emergence of the upper claw and activates the grasping mode. The procedure is the same as that used in a previous methodology using active motor driving. Through this strategy, the passive fingertip rotation creates a space for grasping, as long as there is enough space to insert the fingertip, and the object can slide on a supporting surface.

Finally, the procedure for switching from grasping mode to insertion mode is presented (see Fig. 3). The finger is designed such that switching can be achieved by moving the slide pin or slide base upward. The upward movement results in the connecting pin in the lower link climbing over the protrusion of the lower link. The constraint imposed by the protrusion at the lower link on the motion of the connect pin causes the middle and lower links to transform from the upper claw shape to the original state. Correspondingly, the snapped clip is unlocked. Subsequently, the lower link returns to its original state and the fingertip link returns to its original state in conjunction with the movement of the lower link.

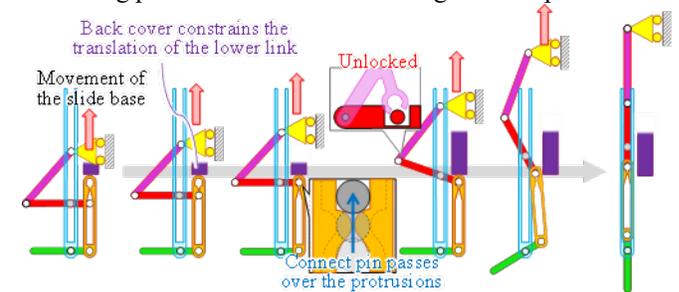

**Fig. 3.** Procedure for switching from the grasping mode to insertion mode

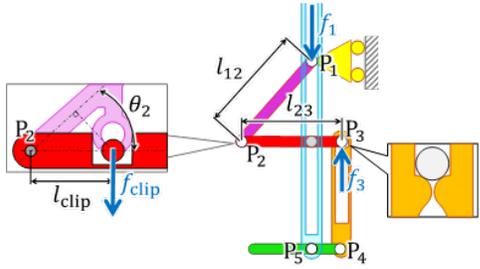

**Fig. 4.** Model of the finger when switching from the insertion mode to grasping mode

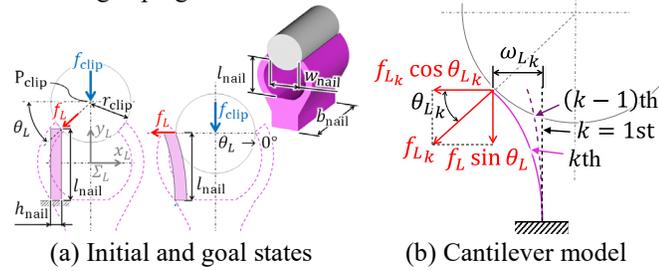

(a) Initial and goal states  (b) Cantilever model

**Fig. 5.** Model used to derive $f_{\text{clip}}^{sn}$

The key components of the robotic finger for mode switching and grasping operations are the protrusion structure (including the spring-clip structure) and parallel links with variable link lengths. The protrusion structure provides discontinuous movement of the links in response to the applied load (the link movement stops unless the load exceeds the threshold for activating the motion). This enables the transformations to occur in the desired order. The transformation provides a parallel link mechanism comprising a guide rail, middle link, lower link, and fingertip link for the grasping operation. In both the lower link and guide rail, the lengths of the parts that serve as the link vary with the movement of the slide and connect the pins. By contrast, the motions in other directions are constrained by the arrangements of the pins or joints in the four-bar link structure, which allows for an antipodal grasping motion.

## III. CONDITION FOR MODE SWITCHING

### A. Analysis

To achieve the transformations in the desired order described in the previous section, a static analysis of the finger is presented in this section. The relationship between the forces applied to the middle link from the spring clip and lower link after the middle link becomes perpendicular to the guide rail was analyzed to determine the order of locking or unlocking the spring clip and protrusion. First, an analysis of the switch from insertion mode to grasping mode is presented. The conditions for mode switching are as follows: 1) The spring clip snaps with the clip pin to couple the upper and middle links before the connect pin passes over the protrusions on the lower link. 2) The power for the mode switching is within the power range of the motor. The nomenclature used for the analysis is shown in Fig. 4. Let $P_i$ ($i \in \{1 \cdots 5\}$) be the points where each link is connected to the other links by the connect pins; $f_{\text{clip}}$ ($\in \mathcal{R}$) be the force applied to the clip pin of the middle link from the spring clip of the upper link; and $f_3$ ($\in \mathcal{R}$) be the force applied to the middle link from the protrusion of the lower link through the connect pin. The moment balance of the middle link around the $P_2$ is

$$l_{\text{clip}} f_{\text{clip}} = l_{23} f_3, \quad (1)$$

where $l_{\text{clip}}$ is the length from the clip pin to $P_2$, and $l_{23}$ is the length of $\overline{P_2 P_3}$. Let $f_3^{i \to g}$ ($\in \mathcal{R}$) be the force required for the connect pin to pass over the protrusions. From (1), the value of $f_{\text{clip}}$ when the pin passes over the protrusions is given by

$$f_{\text{clip}}^{i \to g} = l_{23} f_3^{i \to g} / l_{\text{clip}}. \quad (2)$$

To snap the spring clip with the clip pin before the connect pin passes over the protrusions, the force required for snapping, $f_{\text{clip}}^{sn}$, should satisfy the following condition.

$$f_{\text{clip}}^{sn} < f_{\text{clip}}^{i \to g}. \quad (3)$$

From (2) and (3), the desired order for the transformations is obtained by tuning the shape parameters of the links and protrusions, such that (3) is satisfied. Furthermore, the shape parameters should be determined so that the transformations can be performed at a power level within the power range of the motor. Let $\tau_m$ be the motor torque, $r_{pg}$ be the pitch radius of the pinion gear, $f_1$ ($\in \mathcal{R}$) be the force applied to the upper link from the slide base by the rack-and-pinion mechanism, $l_{12}$ be the length of $\overline{P_1 P_2}$, and $\theta_2$ be the angle between the upper and middle links. Then, $f_1$ and the moment balance at the upper link are expressed as follows:

$$f_1 = \tau_m / r_{pg}, \quad (4)$$
$$l_{\text{clip}} f_{\text{clip}} = l_{12} \cos \theta_2 f_1. \quad (5)$$

Let $\tau_m^{max}$ be the maximum drivable $\tau_m$; then, $f_1$ corresponding to the $\tau_m^{max}$ is derived from (3) and (4). In summary, from (3), (4), and (5), the condition that must be satisfied to achieve the transformations in the desired order for switching from insertion mode to grasping mode is given by

$$f_{\text{clip}}^{sn} < f_{\text{clip}}^{i \to g} = \frac{l_{23} f_3^{i \to g}}{l_{\text{clip}}} < \frac{l_{12} \cos \theta_2 \tau_m^{max}}{l_{\text{clip}} r_{pg}} := f_m^{max}. \quad (6)$$

To evaluate (6) and determine the shape parameters, the magnitudes of $f_{\text{clip}}^{sn}$ and $f_3^{i \to g}$ must be derived. This is achieved by first considering the force for the spring clip ($f_{\text{clip}}^{sn}$). The analytical model is shown in Fig. 5. As shown in Fig. 5(a), the bow-shaped nail (BSN) of the spring clip is assumed to be cantilevered with a concentrated load applied to the tip. Let $f_{\text{clip}}^{sn}$ ($\in \mathcal{R}$) be the force when the BSN, considered as the cantilever beam, is displaced, and the clip pin (gray circle in Fig. 5(a)) passes over the side of the beam. The snapping condition is as follows:

$$r_{\text{clip}} \leq \omega_L + w_{\text{clip}}/2, \quad (7)$$

where $\omega_L$ is the displacement of the tip of the BSN that is considered to be the beam, $r_{\text{clip}}$ is the radius of the clip pin, and $w_\text{clip}$ is the aperture length of the spring clip. Herein, $\omega_L$ caused by the $f^{sn}_\text{clip}$ should satisfy (7). In the field of structural mechanisms, the direction of the force is assumed to be constant. However, in this case, the direction of the force that bends the beam, i.e., the contact force ($f_L$) applied to the BSN from the clip pin, is shifted according to the movement (position) of the clip pin (see Fig. 5(a)). The transition of the mechanical relationship from when the clip pin first contacts the beam (BSN) to the completion of the snapping entails a series of steps. The step at the first contact is defined as $k = 1$st step. The coordinate frame $\Sigma_L$ is located at the center of the aperture of the spring clip, as shown in Fig. 5(a). Let $P_\text{clip}$ be the center point of the clip pin in the $x_L y_L$ plane. We consider the mechanical relationship at the $k$th step, as shown in Fig. 5(b). Let $\theta_{L_k}$ be the angle between the $x_L$ axis and the line to $P_\text{clip}$ from the contact point between the clip pin and the beam (BSN). Let $\omega_{L_k}$ be $\omega_L$ at the $k$ th step. From the geometrical relationship, the difference between $\omega_{L_k}$ and $\omega_{L_{k-1}}$ is

$$d\omega_{L_k} = \omega_{L_k} - \omega_{L_{k-1}} = r_\text{clip}(\cos\theta_{L_k} - \cos\theta_{L_{k-1}}). \tag{8}$$

Assuming that $d\theta_{L_k} := \theta_{L_k} - \theta_{L_{k-1}}$ (the change in $\theta_L$ between the two steps) is sufficiently small, (8) is re-written as

$$d\omega_{L_k} = -r_\text{clip} \sin\theta_{L_k} d\theta_{L_k}. \tag{9}$$

Let $df_{L_k}$ be the force required to bend the beam by $d\omega_k$ in the $k$th step. From the elastic curve equation, $df_{L_k}$ is given by

$$df_{L_k} \cos\theta_{L_k} = \frac{3E_\text{clip}}{l^3_\text{clip}} \frac{b_\text{clip} h^3_\text{clip}}{12} d\omega_{L_k}, \tag{10}$$

where $E_\text{clip}$ is the Young's modulus of the clip and $l_\text{clip}$, $b_\text{clip}$, and $h_\text{clip}$ are the shape dimensions shown in Fig. 5(a), which are the design parameters. $df_L$ is applied to both the left and right beams (BSNs); therefore, $df_\text{clip} = 2df_L \sin\theta_L$. Then, $f^{sn}_\text{clip}$ is given by

$$f^{sn}_\text{clip} = \int_k 2df_{L_k} \sin\theta_{L_k}. \tag{11}$$

Considering that the range of $\theta_L$ is from $\theta_{L_1}$ to $0°$ (see Fig. 5(a)), from (9) and (10), (11) can be re-written as

$$f^{sn}_\text{clip} = \frac{r_\text{clip} E_\text{nail} b_\text{nail} h^3_\text{nail}}{2l^3_\text{nail}} \int_0^{\theta_{L_1}} \sin\theta_L \tan\theta_L \, d\theta_L$$
$$= \frac{r_\text{clip} E_\text{nail} b_\text{nail} h^3_\text{nail}}{2l^3_\text{nail}} \left(\frac{1}{2}\log\frac{1+\sin\theta_{L_1}}{1-\sin\theta_{L_1}} - \sin\theta_{L_1}\right). \tag{12}$$

From the geometrical relationship (Fig. 5(a)), $\theta_{L_1}$ satisfies

$$r_\text{clip} \cos\theta_{L_1} = w_\text{nail}/2. \tag{13}$$

From (12) and (13), the relationship between $f^{sn}_\text{clip}$ and the design parameters, including $b_\text{nail}, l_\text{nail}, h_\text{nail}, w_\text{clip}$ and $r_\text{clip}$, is obtained. If the design parameters are given, $f^{sn}_\text{clip}$ can be derived from (12) and (13). Next, $f^{i\to g}_3$, the force with which the connect pin passes downward over the protrusion to switch from the insertion mode to grasping mode, is derived. This model is illustrated in Fig. 6. When the connect pin passes over the protrusion, it is assumed that the simple supported beams of the lower link bend owing to the contact force applied by the connect pin on the lower link. The procedure for deriving the relationship between $f^{i\to g}_3$ and the shape dimensions of the structures, which are the design parameters, is the same as that used to derive (12) and (13). The coordinate frame $\Sigma_S$, is set as shown in Fig. 6(a), and the analysis is performed on the $x_S y_S$ plane (see Fig. 6(b)). Let $\omega_{S_k}$ be the displacement of the beam at the contact point at the $k$th step and $\theta_{S_k}$ be the angle between the $x_S$ axis and the line from the center point ($P_\text{cp}$) of the connect pin to the contact point. The force $df_{S_k}$ required to generate the displacement $d\omega_{S_k}(= \omega_{S_k} - \omega_{S_{k-1}})$ at the $k$th step is expressed as follows:

$$df_{S_k} \cos\theta_{S_k} = \frac{3E_\text{low} l_\text{low}}{l^2_{\text{low}_1} l^2_{\text{low}_2}} \frac{b_\text{low} h^3_\text{low}}{12} d\omega_{S_k}. \tag{14}$$

where $E_\text{low}$ is Young's modulus of the lower link, and $l_\text{low}$, $l_{\text{low}_1}$, $l_{\text{low}_2}$, $b_\text{low}$, and $h_\text{low}$ are the shape dimensions, as shown in Fig. 6(a). Then, the force $f^{i\to g}_3$ required for the connect pin to pass over the protrusion is given by

$$f^{i\to g}_3 = \int_k 2df_{S_k} \sin\theta_{S_k}$$
$$= \frac{r_\text{cp} E_\text{low} l_\text{low} b_\text{low} h^3_\text{low}}{2l^2_{\text{low}_1} l^2_{\text{low}_2}} \left(\frac{1}{2}\log\frac{1+\sin\theta_{S_1}}{1-\sin\theta_{S_1}} - \sin\theta_{S_1}\right). \tag{15}$$

where $r_\text{cp}$ denotes the radius of the connect pin. From (15), the relationship between $f^{i\to g}_3$ and the design parameters of the lower links is obtained. In summary, the desired transformation sequence for switching from insertion mode to grasping mode, i.e., snapping the pin into the spring clip before the connect pin passes over the protrusion of the lower link, is obtained by setting the design parameters to satisfy (6) using (12) and (15).

Finally, the condition for the desired transformation sequence for switching from grasping mode to insertion mode is derived. The desired order is the reverse of the order when switching from insertion mode to grasping mode: the snapped pin at the spring clip should be unlocked after the connect pin passes upward over the protrusion of the lower link. To realize a different order, the upper and lower areas of the protrusions of the lower links are shaped differently. Consequently, the force needed to unlock the spring clip is the same as the force needed to lock the clip, whereas the force with which the connect pin passed upward and downward over the protrusion of the lower link is different. Let $f^{g\to i}_3$ ($\in \mathcal{R}$) be the force with which the connect pin passes upward over the protrusion to switch from grasping mode to insertion mode. The condition for mode switching is given as follows:

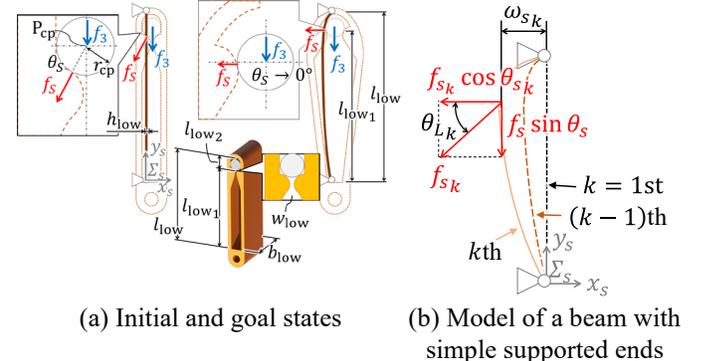

(a) Initial and goal states  (b) Model of a beam with simple supported ends

**Fig. 6.** Model used to derive $f^{i\to g}_3$

$$f_3^{g \to i} < f_{\text{clip}}^{sn} \ (< f_3^{i \to g}). \tag{16}$$

Fig. 7 illustrates the different shapes of the upper and lower parts of the protrusion designed such as to ensure that (16) is satisfied. To reduce $f_3^{g \to i}$, the slope of the lower part is gradual. Similar to the analysis of $f_3^{i \to g}$, $f_3^{g \to i}$ is derived using the theory of a beam with simple supported ends by considering the effect of the gradual slope. In the analysis of $f_3^{i \to g}$, the length $l_{\text{low}_1}$, from the acting point of $f_3^{i \to g}$ to the end of the beam is assumed to be constant; the direction of the force $\theta_S$, on the other hand, varies with the transition of the step (change in contact position). In the analysis of $f_3^{g \to i}$, the slope of the protrusion is gradual; thus, it is assumed that $l_{\text{low}_1}$ varies with $k$ (the number of steps), unlike the acting direction, $\theta_{pr}$, which is constant and perpendicular to the slope of the protrusion, as shown in Fig. 8. Let $l_{\text{low}_{1_k}}$ be $l_{\text{low}_1}$ at the $k$th step. The force $df_{S_k}$ that causes the displacement $d\omega_{S_k}$ at the $k$th step is expressed as follows:

$$df_{S_k} \cos \theta_{pr} = \frac{3 E_{\text{low}} l_{\text{low}}}{l_{\text{low}_{1_k}}^2 \left(l_{\text{low}} - l_{\text{low}_{1_k}}\right)^2} \frac{b_{\text{low}} h_{\text{low}}^3}{12} d\omega_{S_k} \tag{17}$$

Then, the force $f_3^{g \to i}$ required for the connect pin to pass over the protrusion when switching from the grasping mode to the insertion mode is given as follows:

$$f_3^{g \to i} = \int_k 2 df_{S_k} \sin \theta_{pr}. \tag{18}$$

From the geometrical relationship shown in Fig. 8, the displacement $dl_{\text{low}_{1_k}}$, of the connect pin between $k$th and $(k-1)$th steps is expressed as

$$dl_{\text{low}_{1_k}} \tan \theta_{pr} = d\omega_{S_k}. \tag{19}$$

The range of $l_{\text{low}_1}$ is from the $l_{\text{low}_{1_1}}$ to $l_{\text{low}} - l_{\text{low}_2}$, where $l_{\text{low}_{1_1}}$ is the distance from the edge of the slotted hole to the beginning of the slope of the protrusion. From (17), (18), and (19), $f_3^{g \to i}$ is given by

$$f_3^{g \to i} = \int_{l_{\text{low}_{1_1}}}^{l_{\text{low}} - l_{\text{low}_2}} \frac{E_{\text{low}} l_{\text{low}} b_{\text{low}} h_{\text{low}}^3 \tan^2 \theta_{pr}}{2 l_{\text{low}_{1_k}}^2 \left(l_{\text{low}} - l_{\text{low}_{1_k}}\right)^2} dl_{\text{low}_{1_k}}. \tag{20}$$

Note that $l_{\text{low}_2}$ corresponds to $2r_{cp}$ to constrain the movement of the pin during the insertion mode. In summary, the relationship between $f_3^{g \to i}$ and the design parameters of the lower link is derived from (20), and the parameters should be set such that $f_3^{g \to i}$ satisfies (16).

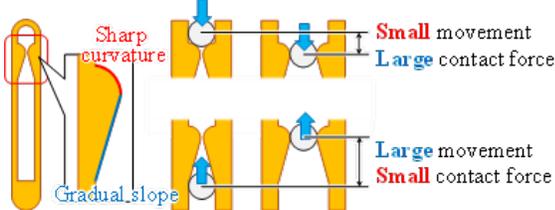

Fig. 7. Shape of the protrusion of lower link

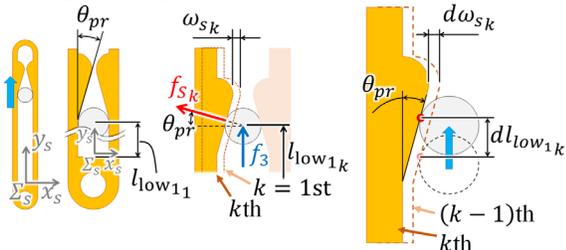

Fig. 8. Model used to derive $f_3^{g \to i}$

### B. Confirmation of analysis

This section confirms the validity of the analysis described in Section III through experiments using a fabricated prototype gripper. The design parameters are listed in Table I. The $f_{\text{clip}}^{sn}, f_3^{i \to g}, f_3^{g \to i}$, and $f_m^{max}$ were measured experimentally using the setups shown in Fig. 9. Additionally, the values of $f_{\text{clip}}^{sn}, f_3^{i \to g}, f_3^{g \to i}$ and $f_m^{max}$ were theoretically derived from (12), (15), and (20), respectively, using the parameters listed in Table I. Finite element method (FEM) analysis (ANSYS Mechanical) was conducted under the same conditions as those in Fig. 9 for confirmation. The obtained values are listed in Table II. As summarized in Table II, the theoretical, FEM, and measured forces were close to each other; thus, the force analysis was validated. Furthermore, the forces satisfied (6) and (16), indicating that the transformations of the prototype gripper were realized in the desired order, as shown in Fig. 1.

## IV. EVALUATION

This section describes an experimental evaluation of the developed gripper. The gripper and manipulator with the attached gripper were manually operated.

### A. Insertable shape of objects

In the finger of the developed gripper, the fingertip was inserted

TABLE I
PARAMETERS OF THE DEVELOPED GRIPPER

| Symbol | Dimension | Symbol | Dimension | Symbol | Dimension |
|---|---|---|---|---|---|
| $l_{\text{clip}}$ | 7.1 mm | $h_{\text{nail}}$ | 0.8 mm | $w_{\text{low}}$ | 1.6 mm |
| $\theta_2$ | 47° | $r_{\text{clip}}$ | 1 mm | $b_{\text{low}}$ | 10 mm |
| $l_{12}$ | 48.6 mm | $l_{\text{low}}$ | 38.6 mm | $h_{\text{low}}$ | 0.9 mm |
| $l_{23}$ | 39 mm | $l_{\text{low}_1}$ | 36.6 mm | $E_{\text{nail}}$ | 2.6 GPa |
| $w_{\text{nail}}$ | 1 mm | $l_{\text{low}_2}$ | 2 mm ($2r_{cp}$) | $E_{\text{low}}$ | 1.4 GPa |
| $l_{\text{nail}}$ | 2.7 mm | $r_{cp}$ | 1 mm | $r_{pg}$ | 10 mm |
| $b_{\text{nail}}$ | 5 mm | $\theta_{pr}$ | 0.3° | $\tau_m^{max}$ | 0.2 Nm |

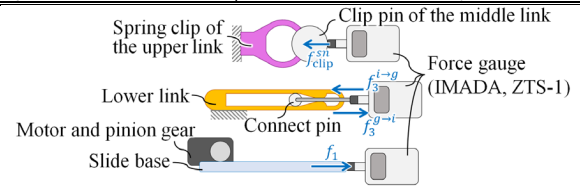

Fig. 9. Experimental setups for measuring $f_{\text{clip}}^{sn}, f_3^{i \to g}, f_3^{g \to i}$ and $f_m^{max}$. The value of $f_m^{max}$ was derived by $f_m^{max} = \frac{l_{12} \cos \theta_2}{l_{\text{clip}}} f_1 \Big|_{\tau_m = \tau_m^{max}}$

TABLE II
THEORETICAL, FEM, AND MEASURED $f_{\text{clip}}^{sn}, f_3^{i \to g}, f_3^{g \to i}$ AND $f_m^{max}$

|  | $f_{\text{clip}}^{sn}$ [N] | $f_3^{i \to g}$ [N] | $f_3^{g \to i}$ [N] | $f_m^{max}$ [N] | Satisfy (6) and (16) |
|---|---|---|---|---|---|
| Theorical | 23.3 | 6.9 | 3.6 | 93.2 | Yes |
| FEM | 25.0 | 7.5 | 4.0 | 116* | Yes |
| Measured | 23.9 | 6.4 | 3.3 | 98.8* | Yes |

* This value was derived from the measured $f_1$ and shape parameters, as listed in Table I (see Fig. 9).

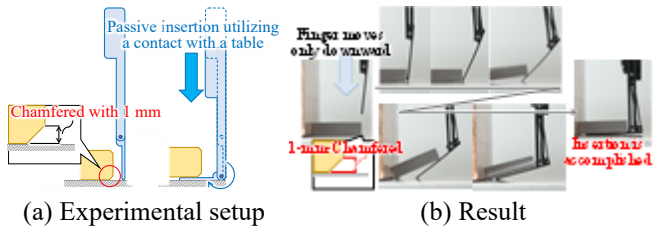

(a) Experimental setup     (b) Result

**Fig. 10.** Insertion test where the fingertip was passively inserted under the object through the chamfered edge

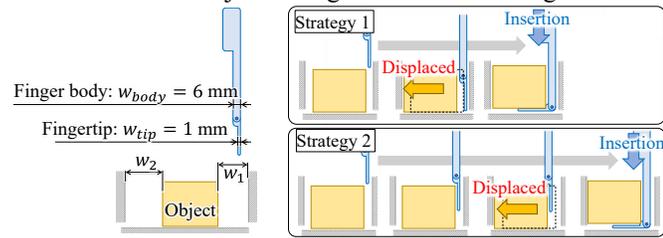

(a) Definition of spaces  (b) Grasping strategy in a narrow space

**Fig. 11.** Grasping an object in a narrow space

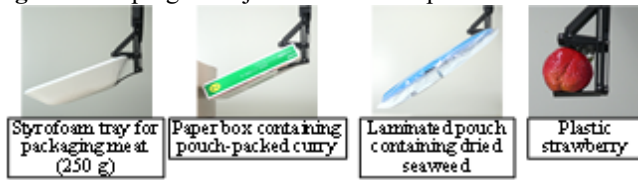

**Fig. 12.** Grasping tests

into the gap between the object and table to scoop and grasp the object. The fingertip was designed to be 1-mm thick; thus, the minimum insertable shape was expected to be a chamfered edge at 1 mm. This was confirmed experimentally using the setup shown in Fig. 10(a). Fig. 10(b) demonstrates that the fingertip could be inserted under the object through a 1-mm chamfered edge.

*B. Grasping in a narrow space*

This section discusses the experiments performed to determine the maximum width of the space between the object and the surroundings that can accommodate a fingertip inserted to grasp the object. The investigation targets were the width of the space, $w_1$, where the fingertip was inserted, and the width of the space, $w_2$, between the object and the surroundings on the opposite side of the inserted space $w_1$, as shown in Fig. 11(a). To obtain the $w_1$ and $w_2$ conditions for insertion, grasping tests were conducted using various values of $w_1$. In the tests, no obstacle was placed on the opposite side of the gripper, with respect to the object, such that the object could freely slide to determine the value of $w_2$ required for grasping. The finger was inserted into the space to grasp the object (Styrofoam tray) using passive fingertip rotation. The observed results in the tests were classified into the following two cases: 1) When $4 \leq w_1$, a finger whose body was than its tip thickness can be inserted into space, and grasping is accomplished through mode switching. Note that because the finger body $w_{body}$ was 6 mm thick, when $(4\leq) w_1 < 6$, the object was displaced by the finger body, and the space for inserting the body expanded as the finger was inserted. This insertion strategy, with passive fingertip rotation and subsequent mode switching, is Strategy 1, as shown in Fig. 11(b). 2) When $1 \leq w_1 < 4$, the finger body cannot be inserted into the space, but the fingertip can be inserted. Therefore, Strategy 2 was adopted, as shown in Fig. 11 (b). After inserting the fingertip into a narrow space, the gripper was moved horizontally to displace the object and enlarge the space for inserting the finger body. Subsequently, the fingertip was inserted under the object, the mode was switched, and the object was successfully grasped. The results indicate that $w_1 + w_2 > w_{body}$ should be satisfied when inserting the fingertip and grasping an object in a narrow space. Insertion was not possible when $w_1 < 1 (= w_{tip})$.

*C. Maximum moment that can be applied to the claw*

As the proposed gripper grasps one end of an object, the moment applied to the base of the claws to hold the object is critical. Subsequently, the maximum moment at the base that can be applied to holding the object was examined. A load was applied to the lower claw during the grasping mode using a force gauge. The acting point of the load was 12 mm from the rotational axis of the fingertip link, i.e., the connect pin. When the applied load reached 18.6 N, slippage occurred between the fingertip link and force gauge owing to the bending of the link. Hence, the maximum applicable moment of the lower claw is 223 Nmm. The applicable moment increased with increase in the thickness of the fingertip; hence, the thickness should be tuned according to the target narrow space and object weight.

*D. Grasping test*

A grasping test was conducted to evaluate the gripper. In the test, the gripper was attached to an automatic positioning stage, and an object placed on a table was grasped. The target objects were as follows: 1) styrofoam tray for packaging meat (250 g), which is commonly used in Japanese supermarkets; 2) rectangular paper box containing pouch-packed curry; 3) flexible and large laminated pouch containing dried seaweed; and 4) complex-shaped plastic strawberry. All the objects with various characteristics were grasped successfully, as shown in Fig. 12. In the tests involving the paper box and strawberry, an obstacle was placed on the opposite side of the finger with respect to the objects to prevent slippage of the objects during insertion (see also the video clip). When slippage occurs during insertion, the approach and obstacle positions should be considered. Moreover, as shown in Fig. 1(b), the claws grasped the object antipodally, similar to a conventional parallel-jaw gripper.

*E. Operation test*

Operational tests were conducted to evaluate the gripper under realistic situations. The evaluation tasks were as follows: 1) picking up an object stuffed in a box and 2) picking up an object in a narrow space partially enclosed at the top and sides. Figs. 1(b) and 1(c) show the results. In the first task, the claws were actively activated and they picked up the target object stuffed in the box. If the second task is performed using a conventional parallel-jaw gripper, the gripper cannot approach the two opposing surfaces of the object owing to the limited workspace. This makes it difficult for the gripper to pick up an object in a narrow space. In contrast, the developed gripper successfully picked up the object by passively inserting its fingertip under the object in the insertion mode and grasping the object while approaching the opposing surfaces in the grasping mode (see the video clip).

## V. CONCLUSION AND DISCUSSION

This paper presented a novel single-fingered reconfigurable robotic gripper with a single actuator for grasping objects in narrow spaces. For this purpose, the gripper is equipped with insertion and grasping modes on its finger. Mode switching is achieved using a single actuator. In insertion mode, the finger is thin, and its fingertip can be rotated actively or passively. The two rotation methods allow for mode switching in two different ways. The low thickness of the finger makes it possible to insert it into a narrow space. The passive rotation of the fingertip enables the insertion of the gripper as it adapts its position to the surroundings. After insertion, the finger configuration is switched from the insertion to grasping mode. Partial passive mode switching made controlling the insertion and grasping simple. However, it was available when a slip occurred between the fingertip and supporting surface, such as a table, and the fingertip rotated on the supporting surface. If a slip does not occur, active mode switching or sophisticated manipulator operations are performed. The developed gripper can grasp objects in narrow working spaces; however, it must grasp one end of an object. The moment applied to the base of the finger to hold the object is critical. A trade-off relationship exists between the applied moment and thickness of the fingertip. The finger was designed to be 1-mm thick to prioritize narrowing of the workable space. If the finger is rotated by 90° after grasping, the moment applied to the base of the finger becomes insignificant. Sophisticated manipulator operations, such as picking up an object through a narrow opening, will be easy if the finger can rotate independently of the movement of the manipulator. The use of multiple robotic arms equipped with the developed grippers can expand the range of tasks that can be accomplished. The implementation of the rotation function and the usage of multiple arms with grippers will be studied in our future work.

During the mode switching, the claws of the mini-gripper were implemented on the structure using a folding mechanism. This paper presents an analysis used to realize the desired transformation sequence for mode switching, and its validity is demonstrated through experiments. As shown in Fig. 1(c), the design that constructs the upper claw after insertion facilitates the grasping of an object in a narrow space with a partially enclosed top. This is particularly effective when the opening area is small. The operation can be performed through a simple control, that is, moving the gripper body only downward. In addition, the adaptability of the developed finger to its surroundings in a narrow space is experimentally verified. The environmental conditions for grasping an object surrounded by obstacles were derived experimentally. The grasping tests demonstrated the grasping ability of the developed gripper. The gripper can grasp commodities handled in supermarkets. In the future, the operation and automation of the developed gripper in narrow spaces will be investigated.